\documentclass{article}
\usepackage{graphicx}
\graphicspath{ {images/} }
\usepackage[export]{adjustbox}
\usepackage{amsmath}
\usepackage{adjustbox}
\usepackage{float}
\usepackage{authblk}
\usepackage{ragged2e}
\usepackage[bottom]{footmisc}
\usepackage[utf8]{inputenc}
\usepackage{url}

\title{Artificial Intelligence in Healthcare: Lost In Translation?}
\author[1,2,3]{Vince I. Madai}
\author[4]{David C. Higgins}

\affil[1]{QUEST-Center for Transforming Biomedical Research, Berlin Institute of Health, Charité Universit{\"a}tsmedizin Berlin, Germany}
\affil[2]{CLAIM - Charité Lab for AI in Medicine, Charité Universit{\"a}tsmedizin Berlin, Germany}
\affil[3]{School of Computing and Digital Technology, Faculty of Computing, Engineering and the Built Environment, Birmingham City University, United Kingdom }
\affil[4]{Berlin Institute of Health, Charité Universit{\"a}tsmedizin Berlin, Germany}

\date{26th of July 2021}

\begin{document}
\maketitle

\begin{center}
\noindent Corresponding Author Details \\
Vince Istvan Madai \\
vince\underline{\space}istvan.madai@charite.de \\
\bigbreak
\noindent Words \\
2955 \\

\end{center}
\newpage
\tableofcontents
\newpage
\begin{abstract}
Artificial intelligence (AI) in healthcare is a potentially revolutionary tool to achieve improved healthcare outcomes while reducing overall health costs. While many exploratory results hit the headlines in recent years there are only few certified and even fewer clinically validated products available in the clinical setting. This is a clear indication of failing translation due to shortcomings of the current approach to AI in healthcare. In this work, we highlight the major areas, where we observe current challenges for translation in AI in healthcare, namely precision medicine, reproducible science, data issues and algorithms, causality, and product development. For each field, we outline possible solutions for these challenges. Our work will lead to improved translation of AI in healthcare products into the clinical setting.
\end{abstract}

\section{Background}
Healthcare systems worldwide suffer from ageing populations and exploding medical costs~\cite{hajat_global_2018}. On the current trajectory, they will soon become economically unsustainable. A possible solution is the application of artificial intelligence (AI) technology, which has the potential to enable improved healthcare outcomes and to reduce overall costs~\cite{deloitte_socio-economic_2020}. 

This can be attributed to the main properties of AI technology. Currently, when we refer to AI, we usually mean techniques of machine learning (ML). ML excels at pattern recognition, especially in big data, and can process a multitude of different input types, ranging from tabular clinical data to medical imaging. These techniques have led to unprecedented advances in areas ranging from computer vision to natural language processing.

But where does AI in healthcare stand today? 

Headlines suggest that AI has already surpassed the performance of human doctors in many medical fields~\cite{longoni_ai_2019,noauthor_ai_2020}. There is no panel on AI in healthcare without the question of when AI will replace human doctors arising. These perspectives stand in stark contrast to clinical reality. Only a few AI tools are available in the clinical setting~\cite{noauthor_medical_nodate,benjamens_state_2020} and we are not aware of a single AI in healthcare tool appearing in clinical guidelines as a standard of care. 

How can this discrepancy be explained? The results gaining media attention are mostly so-called exploratory studies trying to establish proof-of-concepts (PoC) that do not reflect the full process necessary to bring these advancements to clinical reality. Exploratory studies are necessary to identify potentially valuable use cases for AI in healthcare~\cite{kimmelman_distinguishing_2014}. Here, many scenarios need to be tested and successful PoCs can justify - both on an economic as well as an ethical level -  the allocation of funds for follow-up confirmatory studies~\cite{kimmelman_distinguishing_2014}. These confirmatory studies are needed to safely translate the exploratory results to the clinic and need to adhere to all quality markers of studies establishing efficacy, e.g. adequate power and predefined protocols~\cite{kimmelman_distinguishing_2014}. Given the costs of confirmatory studies, they will only be performed as part of product development. And since medical AI products are medical devices, all regulatory and validation requirements of professional medical device development become relevant~\cite{higgins_bit_2020}. 

This path from early-stage PoC to a final medical product is referred to as translation~\cite{sendak_path_2020,beam_translating_2016}. From this point of view, the abundance of exploratory research and the lack of clinically available certified and validated products in the area of AI in healthcare indicates that the translational process needs to be massively improved. 

Importantly, challenges for translation encompass more areas than just product development requirements. In the current perspective, we outline our view of the prerequisites for the successful translation of AI in healthcare products in five categories that we deem as currently most important (see figure 1). We specifically focus on the shortcomings of the status quo and how to improve translation towards the clinical setting.

\begin{figure}[H]
	\includegraphics[width=1\textwidth,center]{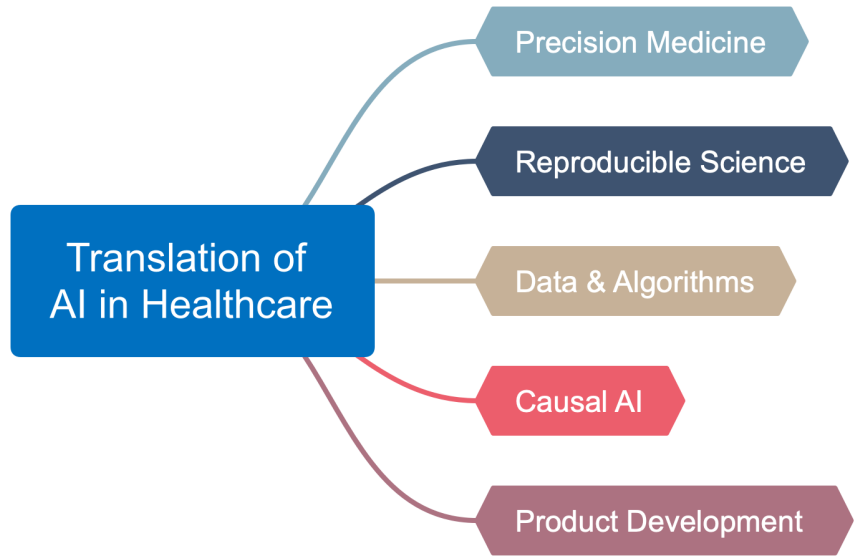}
	\caption{The five main areas where improvements need to happen to facilitate translation of AI in healthcare products.}
\end{figure}

\section{Precision Medicine}
Precision medicine tailors therapies to subgroups of patients rather than to an overall population~\cite{kosorok_precision_2019}. This is advantageous as variation within one patient group can be very large. The average efficacy measured might be driven by a strong efficacy in certain, sometimes small, subgroups neglecting other patient subgroups~\cite{krauss_why_2018}. When one wants to accomplish precision medicine, AI is one of the best-suited methods as it can predict the membership of patients to clinically relevant subgroups~\cite{ronen_evaluation_2019,sinha_machine_2020,khanna_using_2018,piccialli_precision_2021,schork_artificial_2019}. While AI is used for many forms of automation, including in healthcare, it is under-recognised that many important healthcare applications are in fact precision medicine projects~\cite{wilkinson_time_2020,mesko_role_2017,filipp_opportunities_2019,subramanian_precision_2020}.  Many developers are not aware of the situation when their project is actually not only an AI project but also a precision medicine project and underestimate specific challenges threatening translation~\cite{wilkinson_time_2020}. 

If an AI-based precision medicine path is chosen it is important to take into account that confirmatory validation studies for precision medicine are very costly since enough patients per subgroup need to be enrolled. Further, they require expert-level knowledge and experience in clinical validation studies. Projects that do not adhere to such high-quality evidence are bound to fail. It is important to learn the lessons from fields such as cancer research that have adopted precision medicine earlier but suffered from translation failures due to a reliance on low-grade evidence in small subgroups~\cite{kimmelman_paradox_2018}. 

AI-based precision medicine projects are also challenging for translation as their theoretical foundations are currently still shaky since experience is scarce and the methodology is still in its infancy. Differences between and overlaps of the theoretical foundations of AI and statistical inference, estimation and attribution, are still actively researched and important open questions remain~\cite{bzdok_classical_2017,bzdok_inference_2020,efron_prediction_2020}. 

We see a real threat that many AI in healthcare translations will fail due to a lack of understanding of what precision medicine is, and an underestimation of the required level of clinical validation. Thus, there is an urgent need to define which AI healthcare applications are in fact precision medicine approaches and which are not. This conceptual research will aid in decreasing the number of failing projects, streamlining funding to projects with higher chances of success, and educating decision-makers in research and venture funding to be able to critically assess projects presented to them.

\section{Reproducible Science}
Many scientific fields suffer from a reproducibility crisis and this extends to AI in healthcare~\cite{haibe-kains_transparency_2020,hutson_artificial_2018,beam_challenges_2020,carter_pragmatic_2019,mcdermott_reproducibility_2021}. Systematic reviews have shed light on several key factors: Scientific AI publications are frequently under-described due to the lack of reporting standards~\cite{haibe-kains_transparency_2020,liu_comparison_2019}. AI in healthcare models are often tested under unrealistic clinical conditions~\cite{liu_comparison_2019,beede_human-centered_2020}. Due to a narrow focus on homogenous data characteristics, deployment of trained algorithms on new datasets leads to model bias and consequent failures in generalization, i.e. few medical AI systems maintain their performance when confronted with new data and clinical reality~\cite{beede_human-centered_2020,miotto_deep_2018,krois_generalizability_2021}. There is likely also considerable publication bias with cherry-picking of results, e.g. randomly high chance-results are reported~\cite{marin-franch_publication_2018}.

The reproducibility crisis leads to an enormous waste of resources but also to massive impediments to translation~\cite{hauschild_fostering_2021}. The flood of false-positive exploratory studies makes picking the right solutions for translation difficult. Low quality of confirmatory studies will rightfully fail to convince the medical community to adopt AI in healthcare tools. 

To solve this problem lessons can be learned from the approach taken in pharmaceutical drug development that includes reporting standards, documentation, development and maintenance of quality metrics. AI in healthcare reporting standards have been proposed recently, e.g. the MINIMAR guidelines~\cite{hernandez-boussard_minimar_nodate}, others will be available soon like the TRIPOD AI guidelines~\cite{noauthor_tripod_nodate,collins_reporting_2019}. Similarly, guidelines for reporting clinical trial results and for developing clinical trial protocols applying AI in healthcare have been published recently, named the CONSORT-AI and SPIRIT-AI extension, respectively~\cite{liu_reporting_2020,cruz_rivera_guidelines_2020}. Adding a model card~\cite{mitchell_model_2019}and a fact sheet~\cite{arnold_factsheets_2019} to the publication is also encouraged.  

Thorough documentation of every step prevents the loss of pertinent information as a product moves through development. In order to successfully deploy AI in healthcare solutions, honest and repeated testing of the internal and external validity of the models is necessary at all stages in the development process~\cite{higgins_bit_2020}. External validity achieved by applying heterogenous data and appropriate clinical trials is crucial to avoid model bias and overblown clinical performance that will not hold in the clinical setting~\cite{higgins_bit_2020,yu_development_2020,lin_machine-learning_2020}. 

A reduction in the number of false positives among exploratory studies, along with an improvement in the quality of confirmatory studies, will drastically cut the number of futile attempts at translation that are a priori doomed to failure. These improvements will inevitably lead also to improved translation.

\section{Data and Algorithms}
The history of AI is one of capitalizing upon larger and larger corpora of data. Biomedical data, however, suffers from a number of major issues, which were not common among historic machine learning data sets: (i) biomedical data is typically wider than it is tall meaning that dimensionality is high but sample size is relatively small~\cite{efron_prediction_2020}; (ii) sharing, or aggregation, of data sets presents considerable practical and legal difficulties surrounding privacy and anonymization~\cite{abouelmehdi_big_2018}; (iii) medical diagnostic practices change frequently over time, leading to non-stationarity of the data; (iv) the medical signal is often distributed across highly heterogeneous data sets, with many missing entries; (v) bias is often introduced at the data acquisition stage, which leads to subsequent failures in clinical trials. These problems are highly challenging for translation and need tailored solutions.

With regards to the high dimensionality of medical data, false associations can become prevalent. Here, the two-step approach used for genome wide association studies (GWAS) is an excellent example of a structured statistical approach to deal with false associations in wide data~\cite{risch_future_1996}. Other specialised AI methods for high-dimensional data have been proposed~\cite{narita_artificial_2021}. With regards to “not enough data”, federated learning is a much vaunted solution to the problem. In theory, it allows decentralized training of machine learning models on multiple data sets without sharing the underlying data~\cite{rieke_future_2020}. However, when using non-anonymized data - even with federated learning - sensitive information can be at risk~\cite{le_neuraldecipher_2020,kaissis_secure_2020}. AI models have an almost unlimited capacity for recall~\cite{zhang_understanding_2017}. Thus, with current anonymization techniques, a deliberate limitation of their predictive value needs to be performed in order to preserve anonymity~\cite{le_neuraldecipher_2020,li_federated_2020,carlini_secret_2019}. Novel generative AI approaches for anonymization might be the most promising solution to date but still lack the necessary performance~\cite{kossen_synthesizing_2021,fan_survey_2020}.  

Constantly shifting medical standards must be accounted for early in the development process~\cite{challen_artificial_2019}. For a model to be of value it must be able to deliver clinical results for a sufficient amount of time before the clinical standards change again. Importantly, the design of such a clinical solution cannot impede the progress of medical science. This will require a product design focused on developing a solution adapted to the algorithmic realities while keeping a close eye on the medical use case.

Heterogeneous data sets are a prerequisite for AI-powered precision medicine but suffer from specific issues such as missing data that can seriously limit the predictive power of models, even when missing values are imputed~\cite{kossen_framework_2019,poulos_missing_2018}. Most heterogeneous biomedical data come in the form of real-world data (RWD) such as electronic health records. RWD has frequently been posited as a potential solution to the problem of sourcing training data for medical AI~\cite{shah_artificial_2019}.In this case, a relevant ‘signal’ will likely be found for the most prevalent diseases, leading to early and valuable advances. However, the acquisition of sufficient training data becomes exponentially more expensive as disease prevalence diminishes. Additionally, RWD is retrospective by nature lacking prospective choice over which fields are recorded. And more importantly, RWD often lacks accurately recorded outcome labels. With poor control over input fields and a complete lack of quality control in terms of recorded outcomes, it is nigh-on impossible to correctly train a medical AI solution. In order to safely translate an AI-powered solution, built on RWD, to the clinic a thorough validation of the model must therefore be performed. A common source of confusion here derives from comparing machine learning best practices, derived from non-clinical applications, and current expectations for medical device validation and indeed for clinical standards~\cite{higgins_onramp_nodate}. Generally speaking, a validation of the model using only a hold-out set will not be sufficient for the validation of a clinical AI model. Rather a clinical study must be carried out, or similar proxy variables must be found before a model built upon historic data can be deemed to be safe.

Finally, the issue of bias in data acquisition is a core challenge for the translation of AI in healthcare products. Regulatory approval requires the definition of target groups that usually will be patients with a certain disease. In AI variables describing socio-economic status are typically highly predictive of medical history. We do not foresee a regulatory regime that accepts the development of medical treatment models which reinforce existing socio-economic biases. A better approach would be to focus on variables that are biomedical, and not path-dependent, in nature. Here, the development process will have to entail a data acquisition strategy to provide the optimal outcome for all subgroups of patients~\cite{higgins_bit_2020}.

\section{Causal AI}
A major challenge for medical AI systems is to extract ‘true’ signals from this data. This leads to models that initially appear to have phenomenally high-performance, but subsequently fail to generalise to previously unseen data~\cite{krois_generalizability_2021,damour_underspecification_2020}. A further problem is the issue that solutions to biological problems tend to follow not just correlative but highly structured real-world paths. It requires significant technical skill to learn such paths from large data sets, and it is often impossible to learn them from the smaller data-sets more typically seen in bio-medical applications. All of this leads to a high translation failure rate when such models are finally tested in clinical trials. Here, as a solution, causality theory presents a set of techniques which allows boosting of true signal-to-noise ratios, whilst also embedding structure. 

AI, as with epidemiology, is currently in the midst of a causal revolution~\cite{pearl_causality_2009,hernan_causal_2021,rubin_matched_2006,scholkopf_toward_2021}. Typically, machine learning models in AI use either a form of matrix factorization~\cite{rendle_factorization_2010} or cluster-based imputation to generalize hypothetical treatment histories beyond those seen in the training data. Such techniques, due to limitations in the data, lack the statistical power necessary to extract the complex cross-correlations required in medical applications.

New techniques which allow direct assessment of causal effects, through automated methods, obviate the need to execute manual pairwise evaluations of the interactions of tens of thousands of variables. This methodological breakthrough is one of the few realistic hopes which might enable widespread evaluation of RWD for sufficient signal. The adoption of these techniques has been slowed by methodological debates~\cite{pearl_myth_2009,gelman_resolving_2009,king_why_2019}, but is now beginning to gain pace~\cite{sani_identification_2020}, with promising first results~\cite{richens_improving_2020}.

Causal models also have a deeper value than simple prediction and explanation. They can express counterfactuals. In order to perform a medical intervention, the doctor must ask themself, “What if… ?” Causal models are uniquely suited to this purpose due to their position on the “Ladder of Causation”~\cite{pearl_seven_2019}, a mathematical theorem which states that such models can not just predict from examples already seen but can also ‘fill in the table’ of never before seen conjunctions of data points. New intervention policies can be trialed based entirely upon existing historic data, and a firm causal theory, without the need to exhaustively test each scenario in a randomized trial.

Taken together, causality theory, in the form of semi-structured learning approaches, is one of the most promising tools to give a logical underpinning to medical AI solutions and thus critically boost their systematic translation into clinical practice.

\section{Product Development}
Translation of a medical healthcare project into a real-world product requires a focused, market-oriented and professionalized approach to product development. Additionally, appropriate funding for all development stages must be secured. Currently, there is a considerable lack of understanding on the part of all relevant stakeholders - funders, founders and investors - as to how to properly develop and finance AI in healthcare products~\cite{higgins_bit_2020}. This can be attributed to the unique business model of AI in healthcare. 

Despite being a digital technology, medical AI is very different from consumer-tech AI~\cite{sendak_path_2020}. Unfortunately, company founders, in need of raising investment, have muddied the waters on this message because it often suits them to imply that the simplicity of deployment of consumer AI can be married with the high revenue models of healthcare~\cite{higgins_bit_2020}. If funders realise that a business model such as biotech is in fact a better analogy, then they are more likely to switch to funding structures that are more directly aligned with successfully producing quality medical AI products. The major similarity which we observe between AI in healthcare development and biotech is the heavy regulatory burden since a medical AI product falls under the medical device category in the majority of cases.

The obvious solution to the above-mentioned challenges is the standardization of AI in healthcare product development via best practice guidelines and regulatory consolidation. Here, general best practice guidelines outlining the whole process from founding until market launch are needed, as well as specific information on developmental and regulatory processes. With regards to the first, best practice guidelines on AI in healthcare product development have been published recently~\cite{higgins_bit_2020}. Considering specific regulatory guidelines, the USA-based FDA currently leads the pack in terms of regulating medical AI. Their recent action plan~\cite{fda_proposed_2020}, although clearly not yet formal guidelines, demonstrates a willingness to consider even the most advanced medical AI technologies. A formalization into real guidelines can be expected soon. Other budding regulatory guidelines, such as the ones provided by the Johner Institute~\cite{noauthor_johner-institut}, are expected to find uptake from regulatory bodies, especially in Europe. This emerging maturity on the primary regulatory side is matched by further standardization with regards to the development of good machine learning practices (GMLP) and standard-operating-procedures (SOPs) on the technical side~\cite{higgins_onramp_nodate} and structured product development paths~\cite{higgins_bit_2020}. Such a standardized and professionalized approach will give all relevant stakeholders guidance and security for product development and will increase the ratio of successful AI in healthcare translations. 

\section{Conclusion}
We have highlighted key areas where the translation of AI in healthcare products is prone to failure. Additionally, we have outlined promising solutions for the listed challenges. In the pharmaceutical drug development process, several types of key personnel are responsible for numerous aspects in the preparation of what will become the regulatory and validation packages that will ultimately lead to the licensing of a drug for sale. AI in healthcare products will require a comparable level of specialists and professionalized systematic translation processes. In this context, our work will serve as a conversation starter as well as a guide to improve the translation of AI in healthcare products into the clinical setting.

\bibliographystyle{spmpsci}      % mathematics and physical sciences
\bibliography{bibliography}

\newpage
\section{Disclosures}
Dr Madai reports receiving personal fees from ai4medicine outside the presented work. There is no connection, commercial exploitation, transfer or association between the projects of ai4medicine and the presented work.
Dr. Higgins works as senior advisor on AI/ML in healthcare at the Berlin Institute of Health. There is no direct link between results in this article and his income.

\end{document}